\documentclass[lettersize,journal]{IEEEtran}

\usepackage{cite}
\usepackage[colorlinks,linkcolor=blue,citecolor=blue]{hyperref}
\usepackage{graphicx}
\usepackage{epstopdf}
\usepackage{tabularray}
\usepackage{bm}
\usepackage{indentfirst}
\usepackage{float}
\usepackage{CJK}
\usepackage{pifont}

\usepackage{amsmath,amsfonts}
\usepackage{algorithmic}
\usepackage{array}
\usepackage[caption=false,font=normalsize,labelfont=sf,textfont=sf]{subfig}
\usepackage{textcomp}
\usepackage{stfloats}
\usepackage{url}
\usepackage{verbatim}
\usepackage{balance}
\usepackage{multirow}
\usepackage[normalem]{ulem}
\useunder{\uline}{\ul}{}
\usepackage{threeparttable}

\def\BibTeX{{\rm B\kern-.05em{\sc i\kern-.025em b}\kern-.08em
    T\kern-.1667em\lower.7ex\hbox{E}\kern-.125emX}}
\usepackage{balance}
\begin{document}

\title{STNMamba: Mamba-based Spatial-Temporal Normality Learning for Video Anomaly Detection}

\author{Zhangxun Li, Mengyang Zhao, Xuan Yang, Yang Liu, Jiamu Sheng,\\
Xinhua Zeng, Tian Wang,~\IEEEmembership{Senior Member,~IEEE}, Kewei Wu, and Yu-Gang Jiang,~\IEEEmembership{Fellow,~IEEE}\vspace{-10pt}
\thanks{Zhangxun Li, Xuan Yang, Yang Liu, Jiamu Sheng, and Xinhua Zeng are with the Academy for Engineering and Technology, Fudan University, Shanghai 200433, China (e-mail: zhangxunli22@m.fudan.edu.cn; zengxh@fudan.edu.cn). (Corresponding author:
Xinhua Zeng.) 

Mengyang Zhao and Yu-Gang Jiang are with the School of Computer Science, Fudan University, Shanghai, 200433, China.

Tian Wang is with the School of Artificial Intelligence and SKLSDE, Beihang University, Beijing, 100191, China.

Kewei Wu is with the School of Artificial Intelligence, Beijing University of Posts and Telecommunications, Beijing, 100876, China
}}

\maketitle
\begin{abstract}
Video anomaly detection (VAD) has been extensively researched due to its potential for intelligent video systems. However, most existing methods based on CNNs and transformers still suffer from substantial computational burdens and have room for improvement in learning spatial-temporal normality. Recently, Mamba has shown great potential for modeling long-range dependencies with linear complexity, providing an effective solution to the above dilemma. To this end, we propose a lightweight and effective Mamba-based network named STNMamba, which incorporates carefully designed Mamba modules to enhance the learning of spatial-temporal normality. Firstly, we develop a dual-encoder architecture, where the spatial encoder equipped with Multi-Scale Vision Space State Blocks (MS-VSSB) extracts multi-scale appearance features, and the temporal encoder employs Channel-Aware Vision Space State Blocks (CA-VSSB) to capture significant motion patterns. Secondly, a Spatial-Temporal Interaction Module (STIM) is introduced to integrate spatial and temporal information across multiple levels, enabling effective modeling of intrinsic spatial-temporal consistency. Within this module, the Spatial-Temporal Fusion Block (STFB) is proposed to fuse the spatial and temporal features into a unified feature space, and the memory bank is utilized to store spatial-temporal prototypes of normal patterns, restricting the model's ability to represent anomalies. Extensive experiments on three benchmark datasets demonstrate that our STNMamba achieves competitive performance with fewer parameters and lower computational costs than existing methods.
\end{abstract}


\vspace{5pt}

\section{Introduction}
Video surveillance is extensively used in security applications, including public safety, traffic monitoring, and smart homes \cite{r44}. Despite its widespread adoption, identifying abnormal events from lengthy video sequences presents significant challenges. Manual detecting is not only inefficient but also highly susceptible to subjective bias, rendering it inadequate for managing large-scale video data \cite{r1}. Consequently, there is growing interest in the development of intelligent video surveillance systems, with video anomaly detection (VAD) \cite{r3,r4,r5,r6,r7,r8} emerging as a keystone technology.

The primary goal of VAD is to automatically identify anomalies in continuous video frames. However, the rarity and diversity of abnormal events make collecting and labeling large-scale data both costly and challenging, limiting the feasibility of traditional supervised methods reliant on labeled datasets. In this regard, mainstream approaches generally formulate VAD as an unsupervised outlier detection problem. They rely exclusively on normal videos during training to capture regular patterns, with instances deviating significantly from these patterns during testing identified as anomalies. Specifically, prevailing unsupervised methods \cite{r1,r3,r4,r5,r7,r8,r9} typically use proxy tasks to perform VAD by reconstructing the input frame or predicting future frames. The underlying assumption is that models trained solely on regular instances cannot effectively represent anomalies, resulting in notable reconstruction or prediction errors. However, reconstruction-based methods \cite{r1,r3,r4} generally underperform those based on prediction as they tend to overlook temporal information. As a result, prediction-based methods \cite{r7,r8,r9,r38,r45,r56} emerge as the dominant paradigm in unsupervised VAD. 

\begin{figure}[t]
\includegraphics[width=1.0\linewidth]{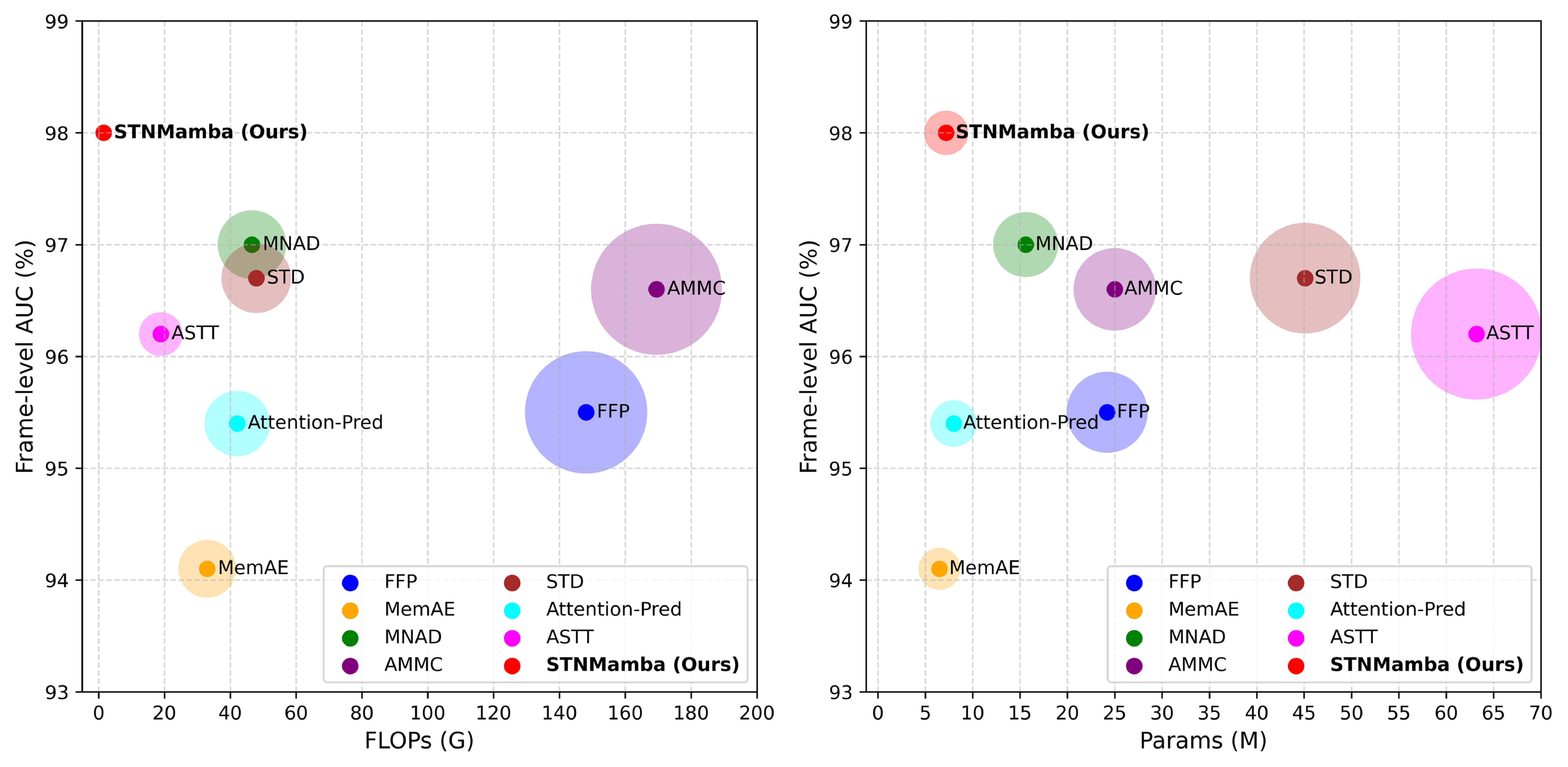}
\centering
\caption{Performance comparisons with respect to FLOPs and Params on the UCSD Ped2 dataset. The size of the circles represents the model's FLOPs or parameters. Our STNMamba outperforms these methods while maintaining notably low model parameters and computational complexity.}
\label{fig1}
\vspace{-15pt} 
\end{figure}

With the advancement of deep learning, previous approaches in VAD depend on convolutional neural networks (CNNs) and transformers. Although CNN-based methods \cite{r5,r7,r8,r9,r24} are widely recognized for their scalability and computational efficiency, the limitations of local receptive fields restrict their ability to learn global contextual information. In contrast, transformer-based methods \cite{r11,r43,r56,r59} can more effectively capture global context and model long-range dependencies, which benefits from the capacity of the self-attention mechanism. This enables transformers to model global spatial relationships and temporal dynamics across video frames, making them better suited than CNNs to handle complex scene variations and semantic changes in video surveillance. However, their self-attention mechanism results in quadratic complexity with respect to the input size, hindering their practical deployment. Recently, Mamba \cite{r12} has garnered considerable attention in language processing tasks due to its global contextual modeling capabilities and linear complexity. Encouraged by this, researchers have introduced Mamba into vision-related domains, inspiring numerous studies \cite{r15,r16,r18}. Naturally, we explore how to effectively leverage Mamba's ability to capture long-range dependencies for enhancing the learning of spatial-temporal normality. However, directly applying Mamba to VAD poses several critical challenges, as outlined below.

First, the vanilla vision Mamba is not fully suitable for spatial-temporal normality learning in VAD tasks. From the spatial perspective, anomalies typically occur in localized regions of an image and often exhibit varying object sizes \cite{r8}. Although VMamba \cite{r16} models the dependencies between 2D image patches using four scanning strategies, it does not explicitly address multi-scale learning, which may lead to missed detections of such anomalies. From the temporal perspective, modeling long-range dependencies with Mamba typically leads to a large number of hidden states in the state space \cite{r54}, which results in redundancy in channel information and may hinder the learning of important temporal dynamics. Therefore, the design of the Mamba block should carefully consider how to effectively extract multi-scale spatial features and capture significant motion patterns, which is pivotal for advancing VAD performance.

\begin{figure*}[ht]
\includegraphics[width=1.0\linewidth]{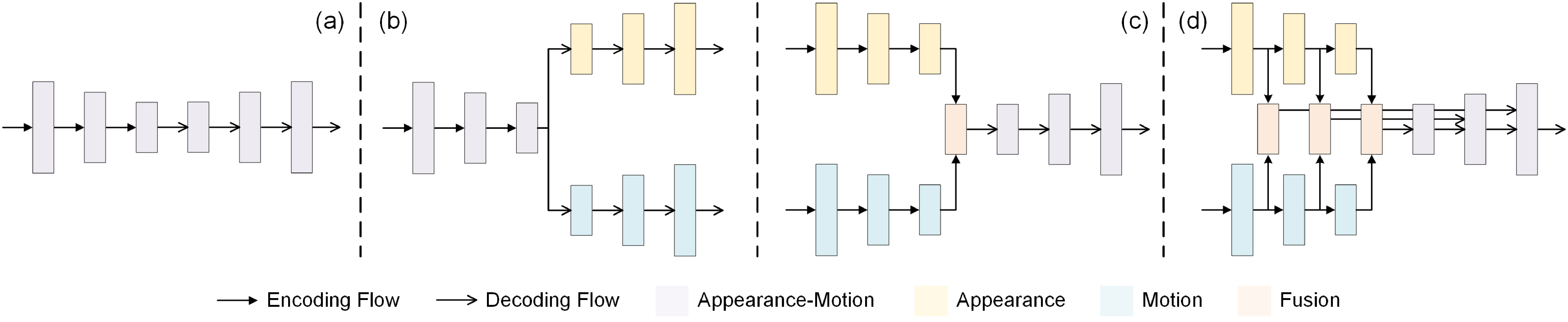}
\centering
\caption{Illustration of the main architectures in unsupervised VAD methods. Unlike existing (a) single-stream networks, (b) dual-stream networks that learn spatial and temporal patterns independently and detect anomalies from the two dimensions, and (c) two-stream networks that perform fusion at the bottleneck, (d) our proposed STNMamba integrates spatial and temporal information seamlessly at multiple levels to model inherent spatial-temporal consistency.}
\label{fig2}
\vspace{-10pt}
\end{figure*}

Second, the architecture of the current VAD methods limits the potential of Mamba in modeling complex spatial-temporal consistency \cite{r19}. To better illustrate this problem, we analyze the prevalent VAD architectures. As illustrated in Fig. \ref{fig2}(a)-(c), they typically develop single-stream \cite{r5,r45,r43,r56} or dual-stream \cite{r7,r8,r9,r38,r58} networks by stacking multiple CNN or transformer blocks to perform encoding and decoding. For single-stream methods, the lack of explicit differentiation between spatial and temporal information hampers feature representation \cite{r9}, thereby limiting overall performance. In contrast, dual-stream methods usually learn spatial and temporal patterns independently, performing fusion \cite{r7,r8,r9} to integrate them or detecting anomalies from the two dimensions separately \cite{r38,r58}. However, all these methods ignore the relationship between spatial and temporal information across multiple levels. This is recognized as key to effectively capturing complex spatial-temporal dependencies in video understanding community \cite{r60, r61}, where shallow features reflect short-term motion dynamics and appearance details, and deep features indicate long-term behavior patterns and semantic information. Such limitations in current architectures hinder their ability to fully leverage Mamba's strength in global sequence modeling, as the limited interaction falls short of modeling comprehensive spatial-temporal consistency. Therefore, designing an architecture that effectively integrates multi-level spatial and temporal features is vital for exploiting Mamba's potential and boosting its performance in VAD tasks.

In view of this, we propose a Mamba-based network (STNMamba) to effectively and efficiently learn robust spatial-temporal normal patterns. To capture the appearance information of objects at different scales, we introduce a Multi-Scale Vision Space State Block (MS-VSSB) in the spatial encoder, which employs parallel depth-wise convolution before the VSSB to extract multi-scale spatial features. Meanwhile, the temporal encoder embeds multiple Channel-Aware Vision Space State Blocks (CA-VSSB) to capture essential temporal dynamics effectively. To model intrinsic spatial-temporal consistency, we design the architecture shown in Fig. \ref{fig2}(d) and develop a Spatial-Temporal Interaction Module (STIM) to seamlessly integrate multi-level spatial and temporal information. Within the STIM, the Spatial-Temporal Fusion Block (STFB) transforms the spatial and temporal features into a unified feature space, and the memory bank stores spatial-temporal prototypes of normal patterns, increasing the gap between normal and abnormal events in feature space. To the best of our knowledge, this is the first work to introduce Mamba to tackle VAD tasks. Experimental results validate that STNMamba achieves competitive performance while maintaining computational efficiency, as illustrated in Fig. \ref{fig1}, making it highly suitable for applications. 

The main contributions of this work are as follows:
\begin{itemize}
  \item We propose a novel Mamba-based network named STNMamba to effectively learn spatial-temporal normality, which is the first work to adapt Mamba to VAD tasks.  
  \item We introduce an MS-VSSB to extract multi-scale spatial features and design a CA-VSSB to capture crucial motion patterns, significantly enhancing the capability of standard Mamba to learn spatial and temporal representations.
  \item We develop an STIM that combines the STFB with the memory bank to seamlessly integrate spatial and temporal information at multiple levels, aiming to exploit Mamba's potential for modeling spatial-temporal consistency.
  \item Extensive experiments on three benchmark datasets demonstrate the superiority and efficiency of STNMamba, achieving competitive performance with remarkably low parameters and computational complexity.
\end{itemize}

\section{Related Work}
\subsection{Unsupervised Video Anomaly Detection}
\subsubsection{Reconstruction-based methods}
In the early stages of VAD research, numerous studies \cite{r1,r3,r4} attempted to reconstruct frames and detect anomalies based on reconstruction errors. They are grounded in the assumption that a model trained only on normal instances is not capable of representing anomalies well, resulting in significant reconstruction errors. For example, Abati \textit{et al.} \cite{r1} incorporated a deep autoencoder with a parametric density estimator, which learns the latent representation of normal patterns through an autoregressive procedure. Fang \textit{et al.} \cite{r3} designed a multi-encoder architecture that extracts motion and content information from observed frames separately. Then, the decoder is responsible for detecting anomalies by measuring the difference between the input frame and the reconstructed frame. However, convolutional neural networks (CNNs) can sometimes generalize well enough to reconstruct anomalies, undermining the detection process. In response, Gong \textit{et al.} \cite{r4} developed a memory network to record prototypical patterns of regular events. They perform the reconstruction task using only a set of normal prototypes stored in the memory bank, which leads to small reconstruction errors for normal samples and larger errors for anomalies. Although these methods made some progress, they do not fully exploit temporal information, thereby limiting overall performance.

\subsubsection{Prediction-based methods}
In contrast to the aforementioned methods, prediction-based methods aim to predict the future frame using consecutive previous frames. They assume that normalities are predictable while anomalies are not. For instance, Liu \textit{et al.} \cite{r5} proposed a future frame prediction framework based on the U-Net and incorporated optical flow as motion constraints. Zhang \textit{et al.} \cite{r6} introduced a GAN-based prediction approach to detect anomalies in both the original image space and the latent space. Nevertheless, these methods only use a sequence of previous frames as input for prediction tasks, ignoring the exploration of spatial and temporal consistency, which leads to sub-optimal performance. Several notable works \cite{r7,r8,r9,r11,r56} recently have been proposed to overcome this limitation. Considering that anomalies can often be distinguished by appearance or motion, some researchers \cite{r7,r8,r9} focused on developing dual-stream frameworks for VAD. These methods typically learn spatial and temporal feature representations separately and then fuse them at the bottleneck of the model. However, they neglect the spatial and temporal features across multi-levels, limiting their ability to effectively model spatial-temporal consistency. To better learn robust spatial-temporal patterns, Yuan \textit{et al.} \cite{r43} introduced a prediction-based method named TransAnomaly, which combines the U-Net with the transformer to capture more comprehensive temporal dynamics and global contextual relationships. Yang \textit{et al.} \cite{r11} proposed a U-shaped Swin transformer network to restore video events, enhancing the extraction of rich dynamic and static motion patterns in videos. Tran \textit{et al.} \cite{r56} introduced a transformer-based network named ASTT for unsupervised traffic anomaly detection. They leverage a transformer encoder to capture spatial and temporal representations and then detect anomalies by prediction error. Although transformer-based methods demonstrate strong capabilities in capturing global dependencies, their computational complexity increases quadratically with respect to image size, causing significant computational overhead. While these methods achieve promising results, they still struggle to comprehensively model spatial-temporal consistency while maintaining a lightweight structure for high efficiency. 

\subsection{State Space Models and Mamba}
Recently, State Space Models (SSMs) \cite{r12,r13}, which originate from classic control theory, have garnered considerable interest among researchers due to their ability to handle computational complexity and efficacy in handling long-range dependency modeling. Specifically, the Structured SSM (S4) \cite{r13} introduced a diagonal structure to address the high computational complexity in previous methods. Subsequently, Mamba \cite{r12} introduced a data-dependent selection mechanism into S4 and proposed an efficient hardware-aware algorithm, ensuring both the effectiveness and efficiency of capturing long-range information. Moreover, the immense potential of Mamba has inspired numerous excellent works \cite{r15,r16,r18} in the realm of vision tasks. For example, Zhu \textit{et al.} \cite{r15} introduced a generic backbone named Vision Mamba (ViM), which incorporates bidirectional sequence modeling for vision tasks. Liu \textit{et al.} \cite{r16} designed a 2D-Selective-Scan (SS2D) module that scans images in the horizontal and vertical directions to capture contextual information. Additionally, Ruan \textit{et al.} \cite{r18} introduced VM-UNet, which incorporates Mamba blocks into a UNet-like architecture to solve medical image segmentation tasks. Nevertheless, the exploration of Mamba in video anomaly detection remains limited. In this study, we leverage Mamba’s capacity to capture global dependencies for modeling the inherent spatial-temporal consistency, offering a novel solution to unsupervised VAD tasks.

\begin{figure*}[ht]
\includegraphics[width=1.0\linewidth]{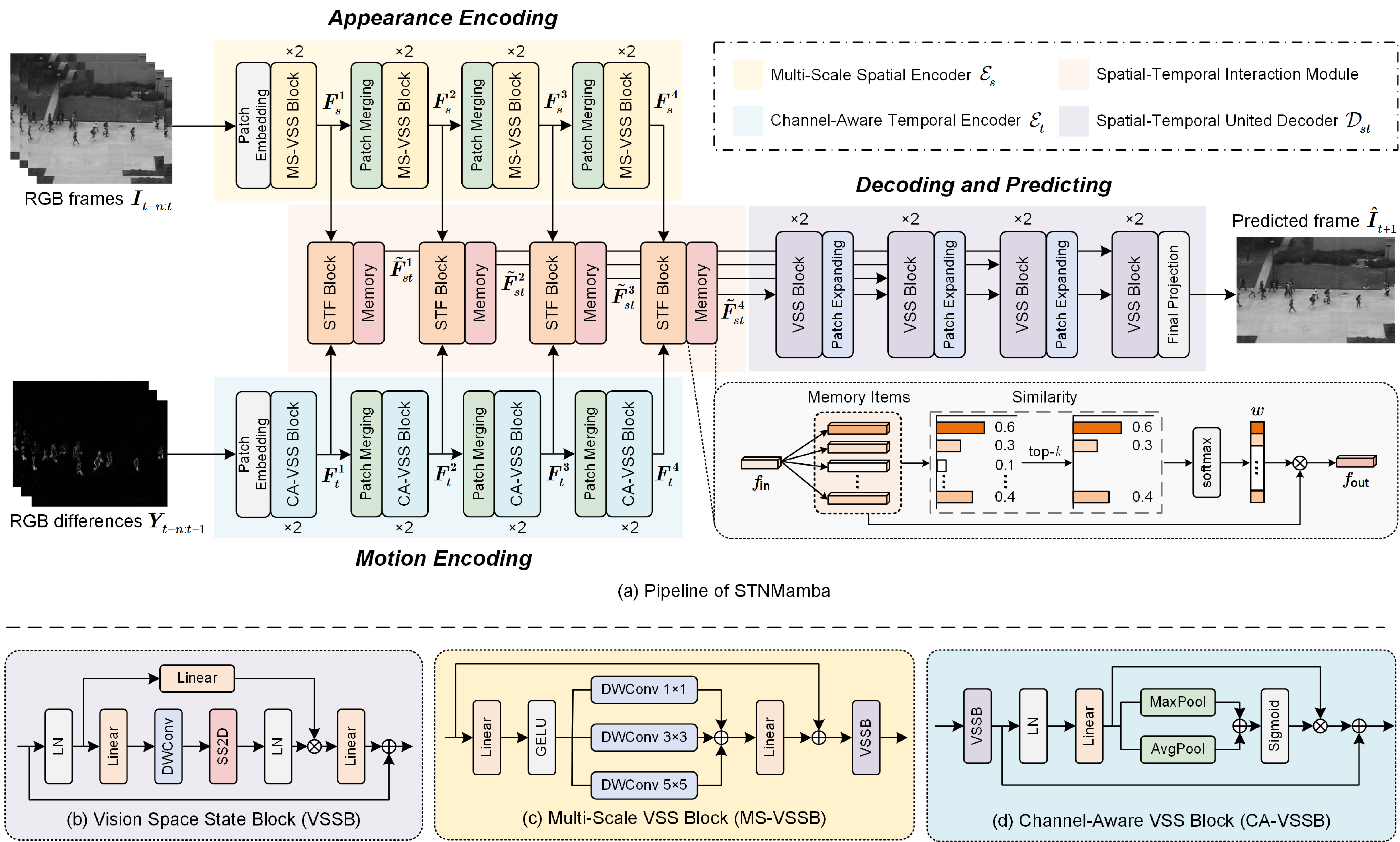}
\centering
\caption{Overview of the proposed STNMamba in (a). The structures of Vision Space State Block, Multi-Scale VSS Block, and Channel-Aware VSS Block are illustrated in (b), (c), and (d), respectively. The proposed STNMamba contains a spatial encoder $\mathcal{E}_{s}$ for appearance encoding, a temporal encoder $\mathcal{E}_{t}$ for motion encoding, a Spatial-Temporal Interaction Module (STIM) for spatial-temporal consistency modeling, and a decoder $\mathcal{D}_{st}$ for decoding and predicting.}
\label{fig3}
\end{figure*}

\section{Method}
\subsection{Architecture Overview}
The overall architecture of the proposed STNMamba is illustrated in Fig. \ref{fig3}(a), consisting of four key components: the Multi-Scale Spatial Encoder $\mathcal{E}_{s}$, the Channel-Aware Temporal Encoder $\mathcal{E}_{t}$, the Spatial-Temporal Interaction Module (STIM), and the Spatial-Temporal United Decoder $\mathcal{D}_{st}$. Firstly, to learn robust spatial-temporal patterns, STNMamba utilizes two carefully designed encoders $\{\mathcal{E}_{s}$, $\mathcal{E}_{t}\}$ based on Visual State Space Blocks (VSSB) to extract multi-scale spatial and crucial temporal features separately. Then, features extracted at each level from the two encoders are fed into the STIM and transformed into a unified feature space, enabling the model to effectively model inherent spatial-temporal consistency. Finally, multi-level spatial-temporal features are processed by the decoder $\mathcal{D}_{st}$, performing decoding and predicting tasks. Compared with the existing dual-stream networks \cite{r7,r8,r9} that typically learn spatial and temporal normality separately and fuse them at the bottleneck of the model, our STNMamba can fully capture multi-level spatial-temporal patterns, allowing it to more effectively detect anomalies.

Specifically, consecutive RGB frames $\boldsymbol{I}_{t-n:t}$ are concatenated along the channel dimension and then fed into $\mathcal{E}_{s}$ for appearance encoding. To fully leverage the semantics of deep features, we propose an MS-VSSB to capture and aggregate multi-scale spatial feature representations. For the motion encoding, considering that the adoption of optical flow results in extra computational overhead, we simply compute the difference of raw RGB values between adjacent video frames to obtain RGB differences $\boldsymbol{Y}_{t-n:t-1}$ as the input of $\mathcal{E}_{t}$. As demonstrated in previous works \cite{r8,r19}, this is an effective way to capture short-term motion patterns and also provide local motion statistics as a complement to the appearance information. Additionally, we introduce a CA-VSSB for the motion encoder, which incorporates channel attention to better capture important motion characteristics. To boost the model’s ability to learn spatial-temporal consistency, multi-level spatial features $\{\boldsymbol{F}_s^1,\boldsymbol{F}_s^2,\boldsymbol{F}_s^3,\boldsymbol{F}_s^4\}$ and temporal features $\{\boldsymbol{F}_t^1,\boldsymbol{F}_t^2,\boldsymbol{F}_t^3,\boldsymbol{F}_t^4\}$ are processed by the STIM, which consists of four STFB and four memory banks to capture unified spatiotemporal representation and store prototypical spatial-temporal patterns. Finally, $\boldsymbol{D}_{st}$ decodes the unified spatial-temporal features $\{\tilde{\boldsymbol{F}}^{1}_{st},\tilde{\boldsymbol{F}}^{2}_{st},\tilde{\boldsymbol{F}}^{3}_{st},\tilde{\boldsymbol{F}}^{4}_{st}\}$ to predict the future frame $\hat{\boldsymbol{I}}_{t+1}$, with anomalies identified by calculating the prediction errors and deviations between the learned memory items during the test phase.

\subsection{Multi-Scale Spatial Encoder}
Mamba has demonstrated superiority in global context modeling for NLP tasks and has achieved initial success in 2D vision tasks \cite{r16}. In contrast to previous works that utilize CNNs \cite{r7, r8} or transformers \cite{r11, r59} to extract appearance features, we propose a Mamba-based encoder to effectively model global spatial dependencies and capture multi-scale feature representations.

For specific implementation, $\mathcal{E}_{s}$ contains four stages to process the input images. Except for the first stage, which consists of a patch embedding layer and two MS-VSSB, the other three stages contain a patch merging layer and two MS-VSSB. Specifically, the patch embedding layer splits the input $\boldsymbol{X}\in\mathbb{R}^{H\times W\times(3\times k)}$ into non-overlapping patches of size $4\times4$ and maps the channel dimension to $C$. The MS-VSSB is designed based on the Vision Space State Block (VSSB) \cite{r16}, which additionally incorporates a set of depth-wise convolution operations using a range of kernel sizes to learn hierarchical feature representations. Furthermore, the patch merging layer is utilized to combine patches along the spatial dimension, which decreases spatial size while increasing the number of channels. After four stages, we obtain multi-level spatial features $\boldsymbol{F}_s^{\{1,2,3,4\}}$, where the channel dimensions for each scale are $[C,2C,4C,8C]$. Details of the VSSB and MS-VSSB are described below.

\subsubsection{Vision Space State Block}
The Vision Space State Block (VSSB) serves as the visual counterpart to Mamba blocks \cite{r12} for modeling long-range dependencies, as shown in Fig. \ref{fig3}(b). Firstly, the input feature $\boldsymbol{X}_{\text{in}}\in\mathbb{R}^{H\times W\times C}$ is processed through two parallel branches. The first branch applies a linear operation followed by a SiLU activation. The second branch involves a sequence of operations: a linear layer, a depth-wise convolution, a SiLU activation, a 2D-Selective-Scan (SS2D) layer, and a Layer Normalization (LN). Subsequently, the outputs from the two branches are then fused using the Hadamard product. Finally, the fused feature is first passed through a linear layer to generate the mixed feature, which is then combined with the input via a residual connection to produce the final output, as follows:
\begin{equation} 
\begin{aligned}
\boldsymbol{X}_{1}&=\sigma(\mathrm{Linear}(\boldsymbol{X}_{\text{in}})),
\\\boldsymbol{X}_{2}&=\mathrm{LN}(\mathrm{SS2D}(\sigma (\mathrm{DWConv}(\mathrm{Linear}(\boldsymbol{X}_\text{in}))))),
\\\boldsymbol{X}_\text{out}&=\mathrm{Linear}(\boldsymbol{X}_{1}\odot \boldsymbol{X}_{2}),
\label{eq1}
\end{aligned}
\end{equation}
where $\boldsymbol{X}_{1},\boldsymbol{X}_{2}\in\mathbb{R}^{H\times W\times \lambda C}$, $\boldsymbol{X}_\text{out}\in\mathbb{R}^{H\times W\times C}$, $\lambda$ represents the expansion factor, $\odot$ denotes the Hadamard product, and $\sigma(\cdot)$ denotes the SILU activation function.

The SS2D is a key component of the VSSB, involving three steps: cross-scan, selective scanning with S6 blocks, and cross-merge. As illustrated in \cite{r16}, the input feature is first flattened into a 1D sequence with the cross-scan along four different directions. Then, four distinctive S6 blocks \cite{r12} are responsible for learning rich feature representations across four directions. Finally, the cross-merge operation sums and merges the sequences to recover the 2D structure.

\subsubsection{Multi-Scale VSS Block}
Since anomalies may occur in regions at different scales in the frame \cite{r8}, we introduce a Multi-Scale VSS Block (MS-VSSB) shown in Fig. \ref{fig3}(c), aiming to effectively extract multi-scale appearance features of objects within the scene. Specifically, a linear projection is first applied to increase the channel dimensions of the input feature. Subsequently, we employ a set of parallel convolutional layers with varying kernel sizes to capture multi-scale appearance information before feeding the feature into the VSSB layer. To reduce parameter overhead and computational costs, we utilize depth-wise convolution, which is well known for its high efficiency. Then, the multi-scale features are fused via element-wise addition and transformed back to the original dimension through another linear projection. Finally, the fused features are passed to the VSSB layer for further processing. The above workflow can be formulated as follows:
\begin{equation} 
\begin{aligned}
\boldsymbol{X}^{'}&=\mathrm{Linear}(\sum_{k\in K}\mathrm{DWConv}_{k\times k}(\sigma (\mathrm{Linear}(\boldsymbol{X}_\text{in})))),
\\\boldsymbol{X}_\text{out}&=\mathrm{VSSB}(\boldsymbol{X_\text{in}}+\boldsymbol{X}^{'}),
\label{eq2}
\end{aligned}
\end{equation}
where $K$ defines a group of parallel convolution kernels with values of $\{1\times1,3\times3,5\times5\}$, and $\sigma(\cdot)$ denotes the GELU activation function.

\subsection{Channel-Aware Temporal Encoder}
It is widely recognized in the VAD community that detecting anomalies depends on both appearance and motion cues \cite{r8}. Most existing two-stream frameworks \cite{r7,r9} use optical flow to capture motion information. However, there are intrinsic differences between optical flow and RGB appearance, and simple fusion strategies such as concatenation or addition may fail to adequately integrate information from these two modalities. Moreover, the use of optical flow can lead to additional computational burden. To address these problems, we utilize the corresponding RGB differences $\boldsymbol{Y}_{t-n:t-1}$ as the input to $\mathcal{E}_{t}$ to effectively capture the motion patterns in context frames, providing a more lightweight and efficient alternative. As shown in Fig. \ref{fig3}(a), the detailed structure of the temporal encoder $\mathcal{E}_{t}$ is similar to that of the spatial encoder $\mathcal{E}_{s}$, except that we replace the MS-VSSB with the proposed Channel-Aware VSS Block (CA-VSSB) to facilitate the learning of significant motion normality. Specifically, as shown in Fig. \ref{fig3}(d), the CA-VSSB incorporates VSSB with channel attention, which consists of the average pooling and max pooling to capture vital temporal dynamics. The workflow of CA-VSSB can be represented as follows:
\begin{equation} 
\begin{aligned}
\boldsymbol{X}_1&=\mathrm{VSSB}(\boldsymbol{X}_\text{in}),\\
\boldsymbol{X}^{'}&=\mathrm{LN}(\mathrm{Linear}(\boldsymbol{X}_1)),\\
\boldsymbol{X}_2&=\boldsymbol{X}^{'}\odot\sigma(\mathrm{AvgPool}(\boldsymbol{X}^{'})+\mathrm{MaxPool}(\boldsymbol{X}^{'})),\\
\boldsymbol{X}_\text{out}&=\boldsymbol{X}_1+\boldsymbol{X}_2,
\label{eq3}
\end{aligned}
\end{equation}
where $\odot$ denotes the Hadamard product, and $\sigma(\cdot)$ denotes the Sigmoid activation function. In this manner, we form a channel-aware scheme for robust motion encoding and obtain the multi-level temporal features $\boldsymbol{F}_t^{\{1,2,3,4\}}$. Experimental results demonstrate that the proposed CA-VSSB can effectively capture important dynamic motion patterns, enhancing the ability of STNMamba to detect anomalies.

\subsection{Spatial-Temporal Interaction Module}
In contrast to previous dual-stream methods \cite{r7,r58}, which separately learn spatial and temporal normality and primarily fuse them using concatenation or addition, we propose a Spatial-Temporal Interaction Module (STIM) that integrates spatial and temporal features across multiple levels to better model inherent spatial-temporal consistency. As depicted in Fig. \ref{fig3}(a), the STIM contains four Spatial-Temporal Fusion Blocks (STFB) and four memory banks to process spatial and temporal features into the unified spatiotemporal representation $\boldsymbol{F}_{st}^{\{1,2,3,4\}}$. Details of the STFB and the memory bank are described as follows.

\begin{figure}[t]
\includegraphics[width=1.0\linewidth]{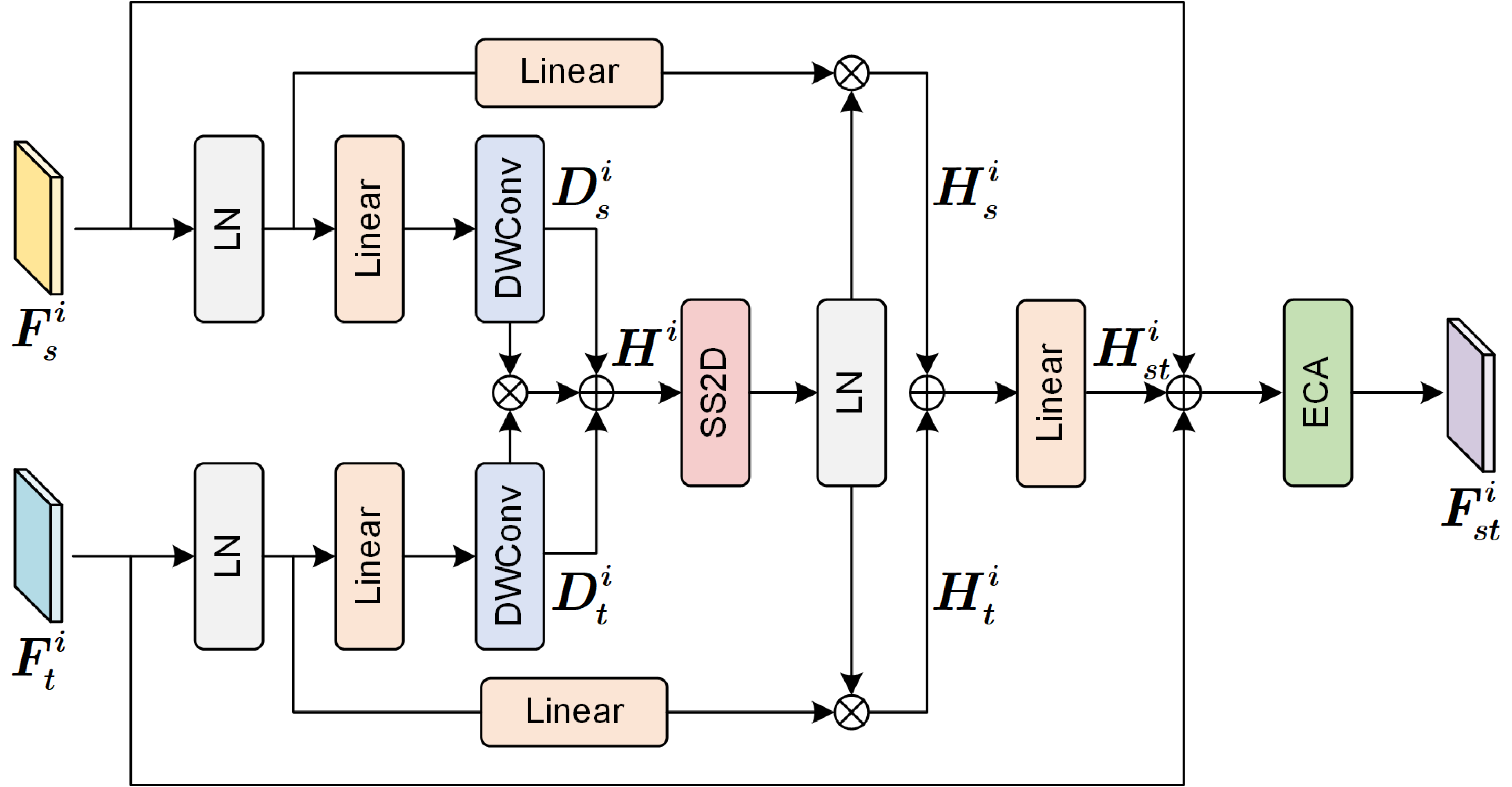}
\centering
\caption{Structure of the proposed Spatial-Temporal Fusion Block (STFB). 
}
\label{fig4}
\vspace{-10pt}
\end{figure}

\subsubsection{Spatial-Temporal Fusion Block}
To effectively explore the intrinsic links between spatial and temporal information, we introduce the STFB to map spatial and temporal features into a unified feature space, boosting the learning of feature representations. As shown in Fig. \ref{fig4}, the spatial feature $\boldsymbol{F}_{s}^{i}$ and temporal feature $\boldsymbol{F}_{t}^{i}$ of $t$th stage are projected into the hidden state space via linear layer and depth-wise convolution layer, and generate the mixed features as follows:
\begin{equation} 
\begin{aligned}
\boldsymbol{D}_{s}^{i}&=\mathrm{DWConv}(\mathrm{Linear}(\mathrm{LN}(\boldsymbol{F}_{s}^{i}))) ,\\
\boldsymbol{D}_{t}^{i}&=\mathrm{DWConv}(\mathrm{Linear}(\mathrm{LN}(\boldsymbol{F}_{t}^{i}))) ,\\
\boldsymbol{H}^{i}&=(\boldsymbol{D}_{s}^{i}\odot \boldsymbol{D}_{t}^{i})+ \boldsymbol{D}_{s}^{i}+ \boldsymbol{D}_{t}^{i}, 
\label{eq4}
\end{aligned}
\end{equation}
where $\odot$ denotes the Hadamard product. Then we fed the hybrid feature $\boldsymbol{H}^{i}$ into an SS2D layer to model long-range dependencies and perform integration with the input features, which can be formulated as follows:
\begin{equation} 
\begin{aligned}
\boldsymbol{H}_{s}^{i}&=\mathrm{LN}(\mathrm{SS2D}(\boldsymbol{H}^{i}))\odot \sigma(\mathrm{Linear}(\mathrm{LN}(\boldsymbol{F}_{s}^{i}))),\\
\boldsymbol{H}_{t}^{i}&=\mathrm{LN}(\mathrm{SS2D}(\boldsymbol{H}^{i}))\odot \sigma(\mathrm{Linear}(\mathrm{LN}(\boldsymbol{F}_{t}^{i}))),
\label{eq5}
\end{aligned}
\end{equation}
where $\sigma(\cdot)$ represents the SILU activation function. Subsequently, the $\boldsymbol{H}_{s}^{i}$ and $\boldsymbol{H}_{t}^{i}$ are projected back to the original space and summed with residual connection from the original input. Finally, the spatial-temporal fused feature $\boldsymbol{F}_{st}^{i}$ is obtained by processing through an ECA layer \cite{r23}, which offers effective channel selection while maintaining efficiency. The process can be represented as follows:
\begin{equation} 
\begin{aligned}
\boldsymbol{H}_{st}^{i}&=\mathrm{Linear}(\boldsymbol{H}_{s}^{i}+\boldsymbol{H}_{t}^{i}),\\
\boldsymbol{F}_{st}^{i}&=\mathrm{ECA}(\boldsymbol{H}_{st}^{i}+\boldsymbol{F}_{s}^{i}+\boldsymbol{F}_{t}^{i}).
\label{eq6}
\end{aligned}
\end{equation}

\subsubsection{Memory Bank}
Memory networks are capable of recording normal patterns to boost the prediction of normal data while suppressing anomalies, which is widely applied in anomaly detection \cite{r4,r7,r24}. To store spatial-temporal prototypes of normal events, we embed the memory bank after each STFB, as depicted in Fig. \ref{fig3}(a). For specific implementation, the memory bank is a learnable two-dimensional matrix with $N$ memory items of dimension $C$ to record prototypical normal patterns, denoted as $\boldsymbol{M}=\{\boldsymbol{m}_1,\cdots,\boldsymbol{m}_N\}\in\mathbb{R}^{N\times C}$. The memory bank involves two key operations: reading and writing. The reading operation intends to reconstruct the input features using normal patterns recorded in the memory items, while the writing operation aims to store prototypical spatial-temporal patterns of normal instances in the memory bank. Both reading and writing operations are conducted during training, whereas only the reading operation is performed during testing, with the parameters of the memory bank remaining fixed.

Specifically, we first expand the spatial-temporal feature $\boldsymbol{F}^{i}_{st} \in \mathbb{R}^{H\times W \times C}$ into query maps along the channel dimension, denoted as $\boldsymbol{Q}=\{\boldsymbol{q}_{1},\cdots,\boldsymbol{q}_{\hat{N}}\} \in \mathbb{R}^{\hat{N}\times C}$, where $\hat{N}=H\times W$. As stated earlier, the reading operation aims to use the memory items $\bm{p}_j$ to reconstruct the query $\boldsymbol{q}_i$ in $\boldsymbol{Q}$ approximately, which can be formulated as follows:
\begin{equation} 
\begin{aligned}
\hat{\boldsymbol{q}_i}=w_i\boldsymbol{M},
\label{eq7}
\end{aligned}
\end{equation}
where $\hat{\boldsymbol{q}_i}$ denotes the $i$th reconstructed query, and $w_i=\{w_i^{(1)},\ldots,w_i^{(j)},\ldots,w_i^{(N)}\}\in\mathbb{R}^{1\times N}$ represents the weighted coefficients obtained by first calculating the cosine similarity between query maps $\boldsymbol{q}_i$ and memory items $\boldsymbol{p}_j$ denoted as $\hat{w_i}$:
\begin{equation} 
\begin{aligned}
\hat{w}_i^{(j)}=\frac{\boldsymbol{q}_i\boldsymbol{p}_{j}^{T}}{\|\boldsymbol{q}_i\|\cdot\|\boldsymbol{p}_{j}\|}.
\label{eq8}
\end{aligned}
\end{equation}

To further enhance the ability of memory items to represent normal patterns, we introduce a top-$k$ scheme to retrieve the top $k\%$ most relevant items in $\boldsymbol{M}$, i.e., we retain the top $k\%$ largest values in $\hat{w_i}$ and set others to $0$. Then we perform a softmax function on $\hat{w_i}$ to calculate the weights $w_i$:
\begin{equation} 
\begin{aligned}
w_i^{(j)}=\frac{\exp(\hat{w}_i^{(j)})}{\sum_{k=1}^N\exp(\hat{w}_i^{(k)})}.
\label{eq9}
\end{aligned}
\end{equation}

Conversely, the writing operation aims to update $\boldsymbol{M}$ using the query features $\boldsymbol{Q}$. Specifically, we first calculate the cosine similarity of $\boldsymbol{m}_j$ to all queries $\{\boldsymbol{q}_{1},\cdots,\boldsymbol{q}_{\hat{N}}\}$ and then perform a softmax function to obtain the weights $\tilde{w}_j$. Subsequently, we adopt the $\mathrm{L}_2$ normalization to ensure that the value scale of updated memory item $\tilde{\boldsymbol{m}}_j$ is the same as the history $\boldsymbol{m}_j$:
\begin{equation} 
\begin{aligned}
\tilde{\boldsymbol{m}}_j=\mathrm{L}_2(\boldsymbol{m}_j+\tilde{w}_j\boldsymbol{Q}).
\label{eq10}
\end{aligned}
\end{equation}

Finally, the original input feature $\boldsymbol{F}^{i}_{st}$ is summed with the retrieved memory feature $\boldsymbol{F}^{i}_{m}$ to obtain the memory-enhanced spatial-temporal feature $\tilde{\boldsymbol{F}}^{i}_{st} \in \mathbb{R}^{H\times W \times C}$:
\begin{equation} 
\begin{aligned}
\tilde{\boldsymbol{F}}^{i}_{st}=\boldsymbol{F}^{i}_{m}+s\cdot \boldsymbol{F}^{i}_{st},
\label{eq11}
\end{aligned}
\end{equation}
where $s\in\mathbb{R}^{C}$ serves as a trainable coefficient to balance the contribution of the original input feature. Compared to previous works that perform concatenation \cite{r9,r24} or simply use the reconstructed features to decode \cite{r4}, this approach not only reduces parameters but also preserves effectiveness.

\subsection{Spatial-Temporal United Decoder}
The decoder performs decoding and prediction tasks by processing the spatial-temporal features at different levels from STIM to generate the future frame $\hat{\boldsymbol{I}}_{t+1}$. Specifically, similar to the encoder, the decoder is organized into four stages, with each stage having channel dimensions of $[8C, 4C, 2C, C]$. In each of the four stages, two Visual State Space Blocks (VSSB) are applied, with a patch expanding layer embedded after each VSSB in the first three stages to upsample the feature maps. The prediction frame is obtained through a final projection layer, which first performs a 4-times upsampling using patch expanding to recover the spatial dimensions, followed by a linear operation to restore channel dimensions.

\subsection{Framework optimization}
To better train the model, we design loss functions that incorporate the prediction loss $\mathcal{L}_{p}$, the feature compactness loss $\mathcal{L}_{c}$ and the feature sparsity loss $\mathcal{L}_{s}$ from four memory banks, weighted by trade-off parameters $\lambda_1$ and $\lambda_2$, as follows:
\begin{equation} 
\begin{aligned}
\mathcal{L}=\mathcal{L}_p+\lambda_1\sum_{i=1}^4\mathcal{L}_c^{i}+\lambda_2\sum_{i=1}^4\mathcal{L}_s^{i}.
\label{eq12}
\end{aligned}
\end{equation}

The prediction loss $\mathcal{L}_{p}$ is defined as the $L_2$ distance between the predicted frame $\hat{\boldsymbol{I}}_{t+1}$ and ground truth $\boldsymbol{I}_{t+1}$, as follows:
\begin{equation} 
\begin{aligned}
\mathcal{L}_{p} = \left\|\hat{\boldsymbol{I}}_{t+1} - \boldsymbol{I}_{t+1}\right\|_2^2.
\label{eq13}
\end{aligned}
\end{equation}

Furthermore, we introduce a feature sparsity loss $\mathcal{L}_{s}$ to encourage diversity among memory bank items, ensuring that they can effectively represent prototypical normal event patterns. Specifically, $\mathcal{L}_{s}$ is calculated by measuring the average $L_2$ distance between the query maps $\boldsymbol{Q}=\{\boldsymbol{q}_{1},\cdots,\boldsymbol{q}_{\hat{N}}\}$ and their two nearest memory items $\{\boldsymbol{m}_i^{1},\boldsymbol{m}_i^{2}\}$, as follows:
\begin{equation} 
\begin{aligned}
\mathcal{L}_{s}=\frac{1}{\hat{N}}\sum_{i=1}^{\hat{N}}\left\|\boldsymbol{q}_i-\boldsymbol{m}_i^1\right\|_2^2-\left\|\boldsymbol{q}_i-\boldsymbol{m}_i^2\right\|_2^2.
\label{eq14}
\end{aligned}
\end{equation}

Conversely, the compactness loss $\mathcal{L}_{c}$ aims to drive the query map $\boldsymbol{q}_i$ and its nearest memory item $\boldsymbol{m}_i^1$ as close as possible to guarantee the representation capability of the memory bank, as follows:
\begin{equation} 
\begin{aligned}
\mathcal{L}_c=\sum_{i=1}^{\hat{N}}\|\boldsymbol{q}_i-\boldsymbol{m}_i^1\|_2^2.
\label{eq15}
\end{aligned}
\end{equation}

\subsection{Anomaly Detection}
We measure the deviation from the trained model in both the image domain and feature space to quantify the extent of anomalies during the testing phase, denoted as $\mathcal{S}_p$ and $\mathcal{S}_d$. Followed by \cite{r5}, we define $\mathcal{S}_p$ as the peak signal-to-noise ratio (PSNR) between the predicted frame $\hat{\boldsymbol{I}}_{t+1}$ and its
corresponding ground truth $\boldsymbol{I}_{t+1}$ as follows, where $N$ is the number of pixels in the video frame:
\begin{equation} 
\begin{aligned}
\mathcal{S}_p=10\log_{10}\frac{\max(\hat{\boldsymbol{I}}_{t+1})}{\frac{1}{N}{\|\hat{\boldsymbol{I}}_{t+1}-\boldsymbol{I}_{t+1}\|_2^2}}.
\label{eq16}
\end{aligned}
\end{equation}

Since the memory bank only stores prototypes of normal patterns, we suggest that query maps derived from anomalous frames result in a significant distance from the memory items. Thus we calculate the $L_2$ distance between $\boldsymbol{q}_i$ and its closest memory item $\boldsymbol{m}_i^1$ to represent $\mathcal{S}_d$, as follows:
\begin{equation} 
\begin{aligned}
\mathcal{S}_d=\frac1{\hat{N}}\sum_{i=1}^{\hat{N}}\|\boldsymbol{q}_i-\boldsymbol{m}_i^1\|_2^2.
\label{eq17}
\end{aligned}
\end{equation}

Subsequently, the anomaly score $\mathcal{S}$ is obtained by integrating the $\mathcal{S}_p$ with $\mathcal{S}_d$, as follows:
\begin{equation} 
\begin{aligned}
\mathcal{S}=\tau (1-g(\mathcal{S}_p))+(1-\tau)g(\sum_{i=1}^4g(\mathcal{S}_d^i)),
\label{eq18}
\end{aligned}
\end{equation}
where $\tau$ denotes the trade-off parameters, $i$ represents the index of the memory bank, and $g(\cdot)$ is the min-max normalization operation to calculate anomaly scores of range $[0, 1]$. 
Finally, we obtain the normality score $\mathcal{N}_s$, as follows:
\begin{equation} 
\begin{aligned}
\mathcal{N}_s=1 -\mathcal{S},
\label{eq19}
\end{aligned}
\end{equation}
where $\mathcal{N}_s$ represents the normal degree of the predicted frames. In other words, a frame with higher $\mathcal{N}_s$ indicates more normal while lower indicates more abnormal.

\section{Experiments}
\subsection{Experimental Setup}
\subsubsection{Datasets}
The following three datasets are used to validate the effectiveness of our STNMamba.

\begin{itemize}
\item \textbf{UCSD Ped2} \cite{r25} contains $16$ training videos and $12$ testing videos with a resolution of $240 \times 360$, which is captured on a pedestrian walkway with an overhead view camera. The dataset defines walking pedestrians as normal, while bicycling, skateboarding, and driving vehicles are regarded as anomalies.
\item \textbf{CUHK Avenue} \cite{r26} contains $16$ training videos and $21$ testing videos with a resolution of $640 \times 360$. The dataset involves $47$ abnormal events such as fast-running, dropping objects, and loitering.
\item \textbf{ShanghaiTech} \cite{r5} contains $330$ training videos and $107$ testing videos with a resolution of $856 \times 480$. This dataset is much more complicated and larger than UCSD Ped2 and CUHK Avenue, which collects $130$ complex and unconventional behaviors such as bicycling and robbing.
\end{itemize}

\subsubsection{Evaluation Metrics}
During the test phase, we can obtain the continuous normality scores of each frame according to Eq. (\ref{eq19}), while the labels of the test frames are discrete. Thus, we calculate the True-Positive Rate (TPR) and False-Positive Rate (FPR) under different thresholds followed by the previous work \cite{r5}. Further, we plot the Receiver Operation Characteristic (ROC) curve and adopt the frame-level average area under the curve (AUC) as our evaluation metrics, with a higher AUC value indicating superior detection performance. 

\subsubsection{Implementation Details}
Our STNMamba is trained on a single Nvidia Geforce RTX 3090 GPU. Adam \cite{r27} is adopted to optimize our model with a batch size of $8$. The learning rate is set to $4\times10^{-{4}}$. All frames are resized to $256 \times 256$ and normalized to the range $[-1, 1]$. The $n$ is set to $4$, indicating we take four consecutive frames as the input. The size of four memory banks, $N_1$, $N_2$, $N_3$ and $N_4$ are empirically set to $80$, $60$, $40$ and $20$, respectively. The trade-off parameters $\tau$, $\lambda_1$ and $\lambda_2$ are set to $0.8$, $0.1$ and $0.01$, respectively.

\vspace{-5pt}

\subsection{Comparison With Existing Methods}
To comprehensively assess the proposed STNMamba, we perform comparative experiments on three benchmarks, comparing our method with SOTA methods in terms of frame-level AUC and efficiency, containing FLOPs, parameters, and FPS.

\begin{table*}[!htbp] 
\captionsetup{skip=0pt} 
\caption{Results of frame-level AUC (\%) and efficiency comparisons with existing methods on three benchmark datasets.}
\label{table1}
\centering
\begin{threeparttable}
\resizebox{1.0\textwidth}{!}{
\renewcommand{\arraystretch}{1.2}
\begin{tabular}{>{\centering\arraybackslash}p{3.5cm}>{\centering\arraybackslash}p{3.5cm}>{\centering\arraybackslash}p{2cm}>{\centering\arraybackslash}p{2cm}>{\centering\arraybackslash}p{2cm}>{\centering\arraybackslash}p{2cm}>{\centering\arraybackslash}p{2cm}>{\centering\arraybackslash}p{2cm}}

\hline
\multirow{2}{*}{Type}  &\multirow{2}{*}{Method} & \multicolumn{3}{c}{Frame-level AUC (\%) $\boldsymbol{\uparrow}$}  & \multirow{2}{*}{FLOPs(G) $\boldsymbol{\downarrow}$} & \multirow{2}{*}{Params(M) $\boldsymbol{\downarrow}$}  & \multirow{2}{*}{FPS(s) $\boldsymbol{\uparrow}$} \\
\cline{3-5}
                                      &                           & UCSD Ped2     & CUHK Avenue   & ShanghaiTech &  &  &  \\
\hline
\multirow{5}{*}{Reconstruction-based} & AbnormalGAN\cite{r36}     & 93.5          & -             & -           & - & - & -\\
                                      & ConvLSTM-AE\cite{r37}     & 88.1          & 77.0          & -           & - & - & -\\
                                      & sRNN-AE\cite{r30}         & 92.2          & 83.5          & 69.6        & - & - & 10  \\
                                      & MemAE\cite{r4}            & 94.1          & 83.3          & 71.2        & 33.0 & \textbf{6.5} & 38 \\
                                      & MESDnet\cite{r3}          & 95.6          & 86.3          & 73.2        & - & - & -  \\
\hline
\multirow{17}{*}{Prediction-based}    & FFP\cite{r5}              & 95.4          & 84.9          & 72.8        & 148.1 & 24.2  & 25 \\
                                      & TransAnomaly\cite{r43}    & 96.4          & 87.0          & -           & - & - & 18\\
                                      & AnoPCN\cite{r29}          & 96.8          & 86.2          & 73.6        & - & - & 10  \\
                                      & MNAD\cite{r24}            & 97.0          & {\uline{88.5}}& 70.5        & 46.6 & 15.6 & \textbf{65}  \\
                                      & Multispace\cite{r6}       & 95.4          & 86.8          & 73.6        & - & - & 18  \\
                                      & AMMC-Net\cite{r7}         & 96.6          & 86.6          & 73.7        & 169.5 & 25.0 & 18  \\
                                      & ST-CAE \cite{r58}         & 92.9          & 83.5          & -           & - & - & -  \\
                                      & STD\cite{r8}              & 96.7          & 87.1          & 73.7         & 47.9 & 45.1  & 32\\
                                      & Attention-Pred\cite{r52}  & 95.4          & 86.0          & 71.4         & 42.1 & 8.0 & 15\\
                                      & STCEN\cite{r39}           & 96.9          & 86.6          & 73.8         & - & - & {\uline{40}}  \\
                                      & MGAN-CL \cite{r51}        & 96.5          & 87.1          & 73.6         & - & - & 30  \\
                                      & ASTNet\cite{r40}          & 97.4          & 86.7          & 73.6         & - & - & -  \\
                                      & MAMC\cite{r9}             & 96.7          & 87.6          & 71.5         & - & - & -  \\
                                      & CL-Net\cite{r42}          & 92.2          & 86.2          & 73.6         & - & - & -  \\
                                      & ASTT \cite{r56}           & 96.2          & 85.4          & -            & {\uline{18.9}} & 63.2 & 35 \\
                                      & PDM-Net\cite{r41}         & {\uline{97.7}}& 88.1          & {\uline{74.2}} & - & - & -  \\
\cline{2-8}
                                      & \textbf{STNMamba (ours)} & \textbf{98.0} & {\textbf{89.0}}& \textbf{74.9} & \textbf{1.5}  & {\uline{7.2}} & {\uline{40}} \\
\hline
\end{tabular}}

\begin{tablenotes}[flushleft]
    \tiny 
    \item \hspace{1em}The bolded and underlined numbers indicate the best and second-best performance, respectively.
\end{tablenotes}
\end{threeparttable}
\end{table*}

\begin{table*}[!htbp] 
\centering
\captionsetup{skip=0pt} 
\caption{Results of component ablation on UCSD Ped2 and CUHK Avenue (\%).}
\label{table2}
\begin{threeparttable}
\renewcommand{\arraystretch}{1.2}
\resizebox{1.0\textwidth}{!}{
\begin{tabular}{>{\centering\arraybackslash}p{1.5cm} >{\centering\arraybackslash}p{2.4cm} >{\centering\arraybackslash}p{2.4cm} >{\centering\arraybackslash}p{2.4cm} >{\centering\arraybackslash}p{2.4cm} >{\centering\arraybackslash}p{2.4cm} >{\centering\arraybackslash}p{0.1cm} >{\centering\arraybackslash}p{2cm} >{\centering\arraybackslash}p{2cm}}
\hline
\multirow{2}{*}{Model} & \multicolumn{5}{c}{Component} & & \multicolumn{2}{c}{Frame-level AUC (\%)} \\
\cline{2-6} \cline{8-9}
                        & MS-VSSB & CA-VSSB & STFB & Memory & Multi-level &   & UCSD Ped2 & CUHK Avenue \\
\hline
1                       & \ding{55} & \ding{55} & \ding{55} & \ding{55} &\ding{55}  &  & 94.3 & 84.9 \\
2                       & \ding{51} & \ding{55} & \ding{55} & \ding{55} &\ding{55}  &  & 94.9 & 85.5 \\
3                       & \ding{51} & \ding{51} & \ding{55} & \ding{55} &\ding{55}  &  & 95.2 & 86.1 \\
4                       & \ding{51} & \ding{51} & \ding{51} & \ding{55} &\ding{55}  &  & 96.6 & 87.6 \\
5                       & \ding{51} & \ding{51} & \ding{51} & \ding{51} &\ding{55}  &  & 97.1 & 88.2 \\
6                       & \ding{51} & \ding{51} & \ding{51} & \ding{51} &\ding{51}  &  & \textbf{98.0} & \textbf{89.0} \\
\hline
\end{tabular}}

\begin{tablenotes}[flushleft]
    \tiny 
    \item \hspace{1em}The bolded numbers indicate the best performance.
\end{tablenotes}
\end{threeparttable}

\end{table*}

\subsubsection{Frame-Level AUC} According to the mainstream methods in VAD, the approaches involved in our comparison experiments can be divided into reconstruction-based methods \cite{r3,r4,r30,r36} and prediction-based methods \cite{r8,r9,r43,r41}. The comparison results are depicted in Table \ref{table1}. Prediction-based methods are usually more effective in leveraging both spatial and temporal information, thus generally outperforming reconstruction-based methods, as shown in Table \ref{table1}. Remarkably, our proposed STNMamba achieves state-of-the-art performance on three benchmark datasets, yielding frame-level AUCs of $98.0\%$, $89.0\%$, and $74.9\%$, respectively. Compared to the second-best methods, STNMamba delivers notable improvements, achieving AUC gains of $0.3\%$, $0.5\%$, and $0.7\%$ on UCSD Ped2 and CUHK Avenue and ShanghaiTech, respectively. These results demonstrate that our STNMamba can more effectively learn spatial and temporal normality, showing superiority in detecting anomalies. In particular, our STNMamba performs best on the ShanghaiTech dataset, which is a complicated multi-scene dataset and has proved to be a challenge for most unsupervised VAD methods \cite{r44}. This success can be attributed to two key factors. First, STNMamba employs two carefully designed Mamba-based encoders that are equipped with the proposed MS-VSSB and CA-VSSB to facilitate the learning of spatial and temporal patterns. Second, the developed STIM enhances the model's capability to capture multi-level spatial-temporal patterns, enabling it to model inherent spatial-temporal consistency and thereby boost overall performance. Consequently, STNMamba demonstrates significant advantages in tackling VAD tasks, particularly in complex scenarios.

\subsubsection{Model Complexity and Efficiency} To showcase the potential of our method for deployment on resource-limited devices, we conduct a quantitative evaluation of both detection performance and model efficiency using the UCSD Ped2 dataset. Although many existing methods do not report their model parameters and inference speed, we attempt to collect these data by gathering the reported results and reproducing others using their open-source code. For a fair comparison, all models are evaluated on an Nvidia GTX TITAN XP GPU. The results are illustrated in Table. \ref{table1}. Notably, compared to MemAE \cite{r4}, our STNMamba achieves a significant 3.9\% performance improvement with only a minimal increase of 0.7M in model parameters. Compared to the transformer-based method ASTT \cite{r6}, our STNMamba outperforms it with fewer parameters and fewer FLOPs. Moreover, STNMamba achieves a 1.0\% improvement with only 1/2 of the parameters and 1/40 of the flops of MNAD, further validating its effectiveness in lightweight model design while achieving superior performance. In terms of the average inference speed, our STNMamba achieves 40FPS, which means it takes an average of 0.025 seconds to determine the anomalies in an image, meeting the demand for real-time detection. Although MNAD \cite{r24} achieves slightly faster inference, their detecting performance is far inferior to our method. These experimental results demonstrate that STNMamba achieves the best trade-off between performance and efficiency, highlighting its strong potential for deployment on resource-constrained edge devices.

\subsection{Ablation Study}
\subsubsection{Effectiveness of key components}
To verify the contribution of each component within the proposed STNMamba, we perform an ablation analysis to observe performance changes, as shown in Table \ref{table2}. Model 1 serves as the baseline, where the Vision Space State Block (VSSB) is used in both the spatial and temporal encoders. This model learns spatial and temporal features independently and then integrates them at the bottleneck using concatenation, yielding an AUC performance of $94.3\%$ on UCSD Ped2 and $84.9\%$ on CUHK Avenue. Next, we equip the spatial encoder with MS-VSSB (Model 2), resulting in additional performance gains of $0.6\%$ on UCSD Ped2 and $0.5\%$ on CUHK Avenue, demonstrating the importance of capturing multi-scale appearance information for detecting local anomalies. Furthermore, Model 3 experiences a $0.6\%$ AUC gain on the CUHK Avenue dataset compared to Model 2, illustrating that capturing important motion patterns is vital for boosting performance. Model 4 demonstrates that the proposed STFB can effectively enhance the model's capability to model inherent spatial-temporal consistency, achieving noticeable AUC performance gains of $1.4\%$ and $1.5\%$ on UCSD Ped2 and CUHK Avenue datasets. With the integration of the memory bank, Model 5 achieves a $0.6\%$ AUC gain on the CUHK Avenue dataset compared to Model 4. This result highlights the efficacy of the memory bank in storing representative spatial-temporal prototypes, thereby enhancing overall performance. In addition, our multi-level fusion strategy encourages the STNMamba to learn robust spatial-temporal features from different levels. Compared to Model 5, which only fuses at the bottleneck, Model 6 brings additional performance gains of $0.9\%$ and $0.8\%$ on UCSD Ped2 and CUHK Avenue datasets, further demonstrating the significance of multi-level spatial-temporal information for VAD tasks.

\subsubsection{Impact of memory banks}
To further illustrate the effectiveness of the four memory banks in our model, we progressively remove them and analyze the performance of the resulting model variants, as depicted in Table \ref{table3}. Specifically, $\mathcal{M}_4$ denotes the memory bank embedded in the bottleneck of the model, while $\mathcal{M}_1$ refers to the memory bank located at the first level of the encoder. As depicted in Table \ref{table3}, removing the memory bank at different stages leads to a performance drop. Specifically, removing $\mathcal{M}_4$ causes the largest decrease, resulting in a $0.6\%$ AUC degradation. Model 5 is a memory-free model variant that does not utilize the memory bank to record normal patterns, suffering obvious performance degradation. The results further suggest that hierarchical memory banks are beneficial for capturing rich spatial-temporal prototypes, thereby enhancing detection performance.

\subsubsection{Effect of loss functions}
We assess different combinations of loss functions to verify their effectiveness in optimizing the model, as presented in Table \ref{table4}. Model 1 is trained solely by minimizing the $L_2$ distance between the predicted frame and the ground truth, achieving only a $87.7\%$ frame-level AUC performance on the CUHK Avenue dataset. In addition, Model 2 achieves the second-best performance, indicating that the compactness loss $\mathcal{L}_{c}$ is more crucial than the feature sparsity loss $\mathcal{L}_{s}$. Model 4 shows that the combination of $\mathcal{L}_{c}$ and $\mathcal{L}_{s}$ can further strengthen the memory bank's ability to represent normal events, achieving an AUC gain of $1.3\%$ compared to Model 1.

\begin{table}[!htbp]
\centering
\captionsetup{skip=3pt} 
\caption{Results of memory bank ablation on CUHK Avenue (\%).}
\label{table3}
\begin{threeparttable}
\renewcommand{\arraystretch}{1.2}
\resizebox{1.0\columnwidth}{!}{
\begin{tabular}{>{\centering\arraybackslash}p{1.5cm} >{\centering\arraybackslash}p{1.125cm} >{\centering\arraybackslash}p{1.125cm} >{\centering\arraybackslash}p{1.125cm} >{\centering\arraybackslash}p{1.125cm} >{\centering\arraybackslash}p{2cm}}
\hline
\multirow{2}{*}{Model}  & \multicolumn{4}{c}{Memory bank}     & \multirow{2}{*}{CUHK Avenue} \\
\cline{2-5} 
                         & $\mathcal{M}_1$    & $\mathcal{M}_2$   & $\mathcal{M}_3$   & $\mathcal{M}_4$   &    \\
\hline                    
1                        & \ding{51}          & \ding{51}         & \ding{51}         & \ding{51}         & \textbf{89.0}  \\
2                        & \ding{51}          & \ding{51}         & \ding{51}         & \ding{55}         & 88.5  \\
3                        & \ding{51}          & \ding{51}         & \ding{55}         & \ding{55}         & 88.3  \\
4                        & \ding{51}          & \ding{55}         & \ding{55}         & \ding{55}         & 88.1  \\
5                        & \ding{55}          & \ding{55}         & \ding{55}         & \ding{55}         & 88.0  \\
\hline
\end{tabular}}
\begin{tablenotes}[flushleft]
    \tiny 
    \item \hspace{1em}The bolded numbers indicate the best performance.
\end{tablenotes}
\end{threeparttable}
\end{table}

\begin{table}[!htbp]
\centering
\captionsetup{skip=3pt} 
\caption{Results of training losses ablation on CUHK Avenue (\%).}
\label{table4}
\begin{threeparttable}
\renewcommand{\arraystretch}{1.2}
\resizebox{\columnwidth}{!}{
\begin{tabular}{>{\centering\arraybackslash}p{1.5cm} >{\centering\arraybackslash}p{1.5cm} >{\centering\arraybackslash}p{1.5cm} >{\centering\arraybackslash}p{1.5cm} >{\centering\arraybackslash}p{2cm}}
\hline
\multirow{2}{*}{Model}  & \multicolumn{3}{c}{Loss function}     & \multirow{2}{*}{CUHK Avenue} \\
\cline{2-4}
                         & $\mathcal{L}_p$    & $\mathcal{L}_c$   & $\mathcal{L}_s$   &    \\
\hline                         
1  & \ding{51}          & \ding{55}         & \ding{55}                               & 87.7  \\
2  & \ding{51}          & \ding{51}         & \ding{55}                               & 88.5  \\
3  & \ding{51}          & \ding{55}         & \ding{51}                               & 87.9  \\
4  & \ding{51}          & \ding{51}         & \ding{51}                               & \textbf{89.0}     \\
\hline
\end{tabular}}
\begin{tablenotes}[flushleft]
    \tiny 
    \item \hspace{1em}The bolded numbers indicate the best performance.
\end{tablenotes}
\end{threeparttable}
\end{table}

\subsubsection{Sensitivity to hyperparameters $\tau$ and $k$}
We quantitatively analyze the sensitivity to key hyperparameters $\tau$ and $k$ on the UCSD Ped2 and CUHK Avenue datasets, as depicted in Fig. \ref{fig5}. As previously stated, the $\tau$ denotes the score weight parameter, which is primarily accountable for balancing the PSNR $\mathcal{S}_p$ and feature distance $\mathcal{S}_d$. By examining the two dashed lines, we observe that the AUC initially rises and then falls. The best AUC performance of $98.0\%$ and $89.0\%$ on the UCSD Ped2 and CUHK Avenue datasets is obtained when $\tau$ is set to 0.8. In addition, we explore the sensitivity to $k$, which represents the percentage of selected memory items during the reading operation. A too-small $k$ may lead to information loss, while a too-large $k$ may fail to retrieve representative prototype patterns, thereby affecting the performance. As illustrated in Fig. \ref{fig5}, we obtain the best performance when $k$ is set to 60, which means that we retrieve the top $60\%$ most relevant items in the memory bank during the reading operation.

\subsection{Visualization Analysis}
\begin{figure*}[!htp]
    \centering
    \begin{minipage}{\textwidth}
        \includegraphics[width=0.8\linewidth]{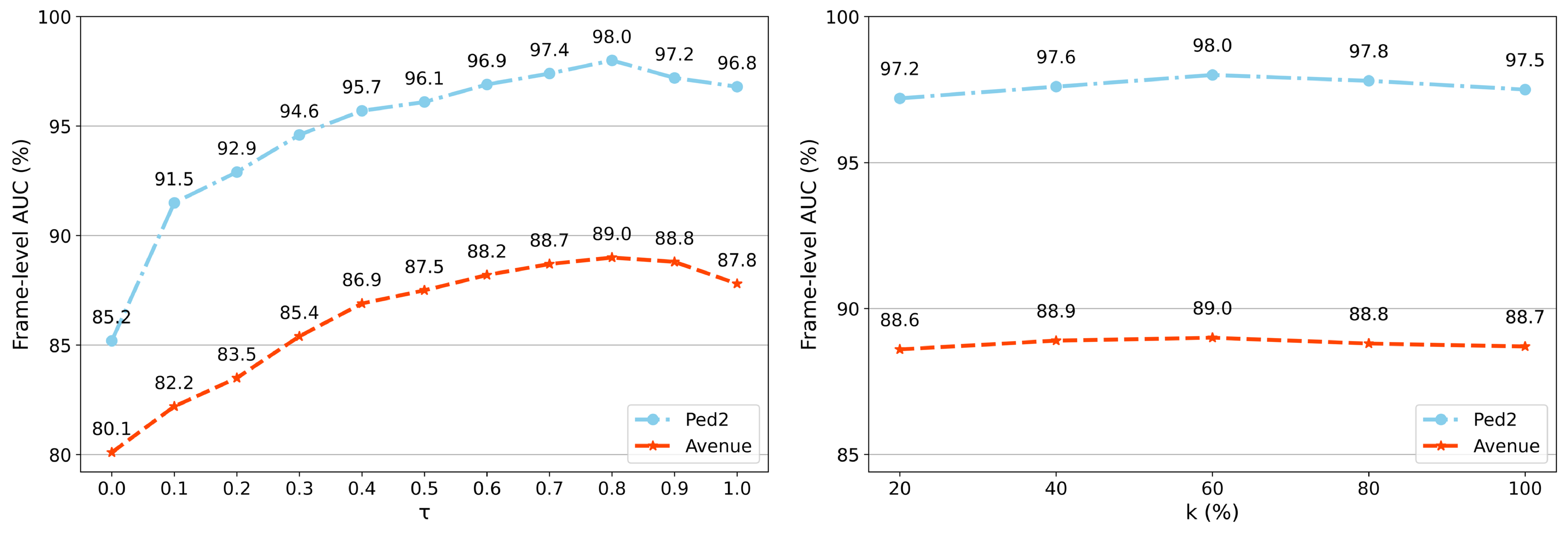}
        \centering
        \captionsetup{skip=3pt} 
        \caption{Results of sensitivity analysis to hyperparameters $\tau$ (left) and $k$ (right) on the UCSD Ped2 and CUHK Avenue datasets.}
        \label{fig5}
    \end{minipage}
    \\[0.5em] 
    \begin{minipage}{\textwidth}
        \includegraphics[width=0.80\linewidth]{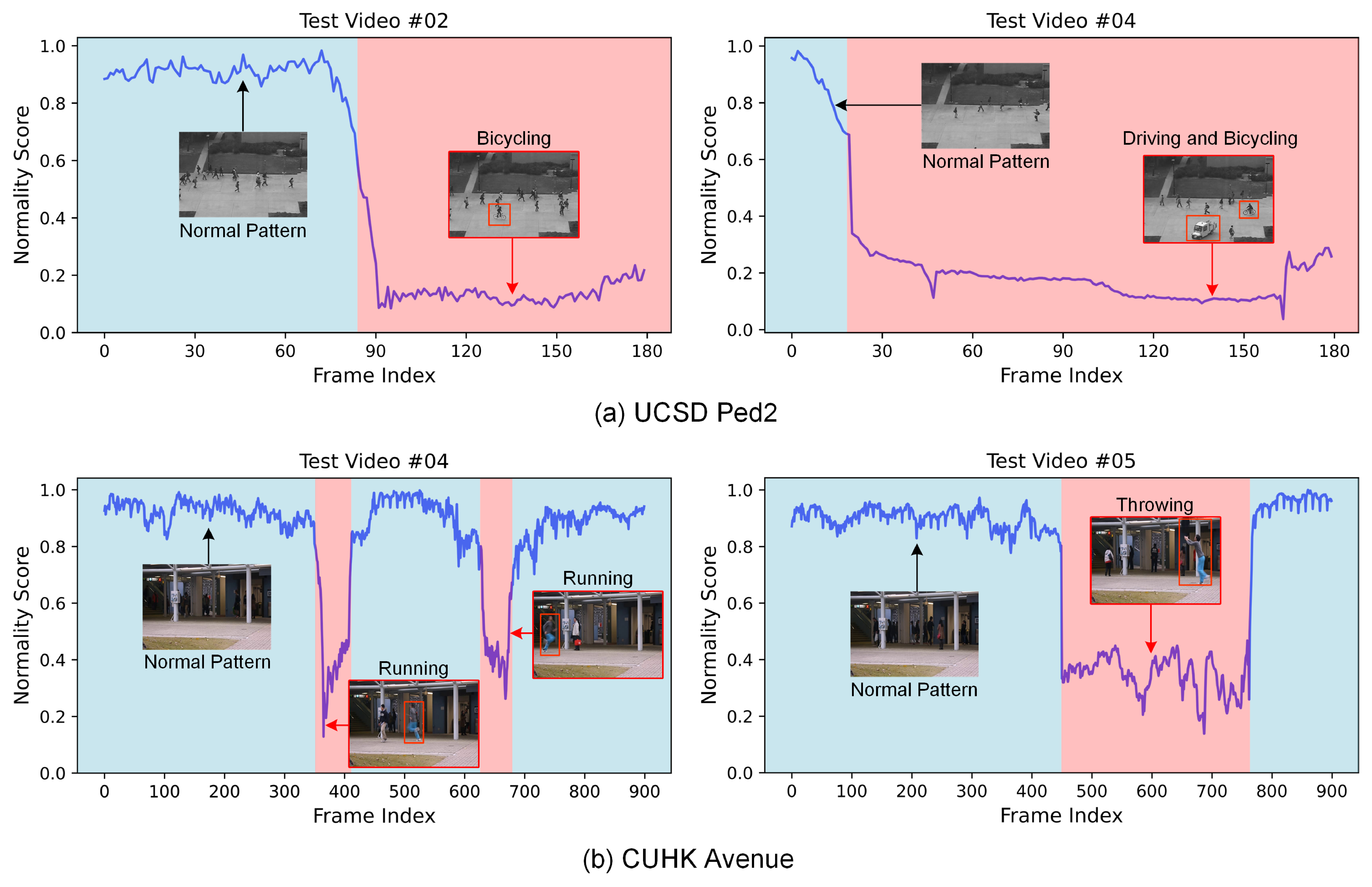}
        \centering
        \captionsetup{skip=3pt} 
        \caption{The normality score curves of some test video clips on the UCSD Ped2 (a) and CUHK Avenue (b) datasets. The red window indicates the time interval for anomalous frames while the blue window represents normal frames.}
        \label{fig6}
    \end{minipage}
    \\[0.5em] 
    \begin{minipage}{\textwidth}
        \includegraphics[width=0.80\linewidth]{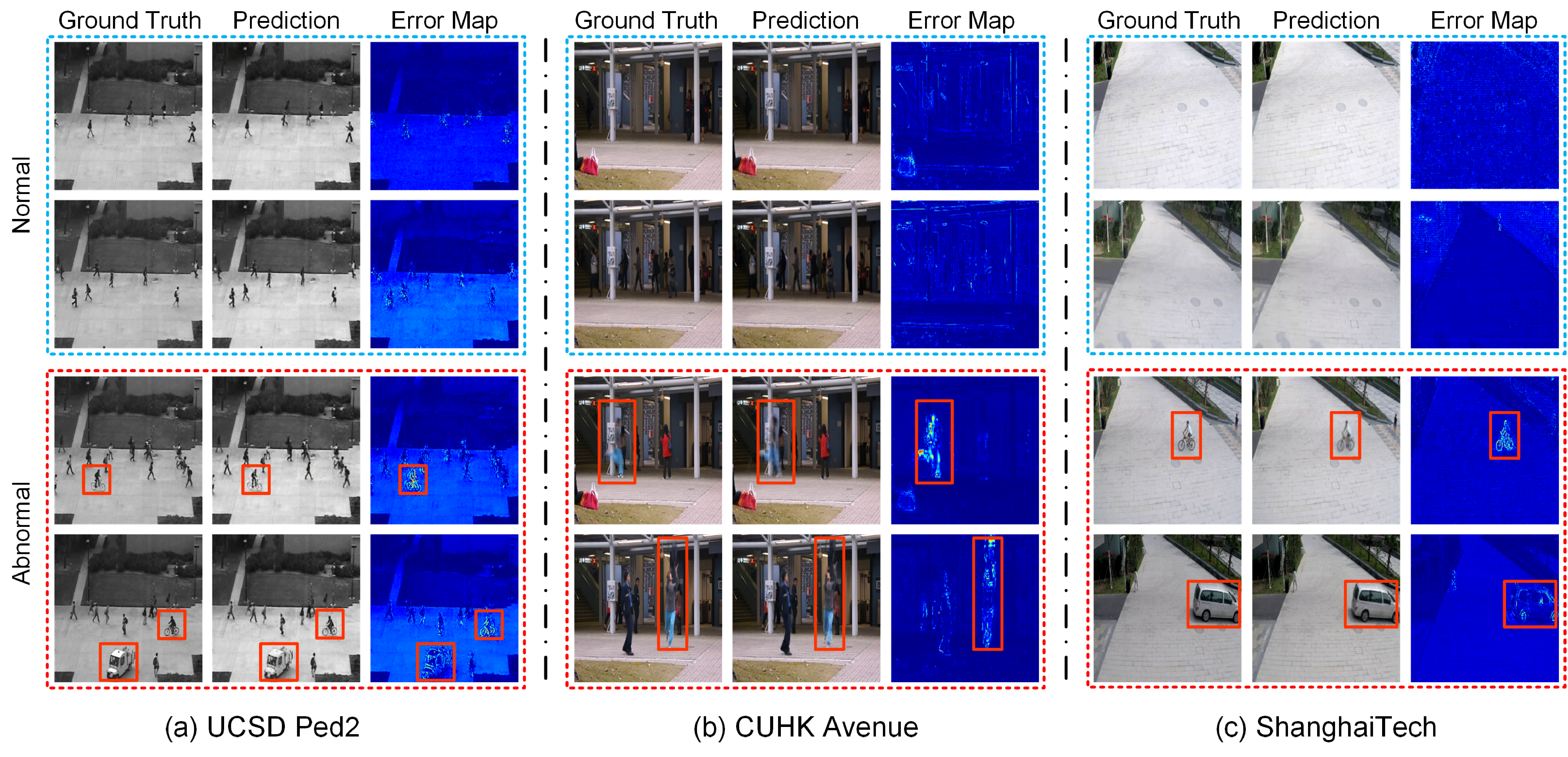}
        \centering
        \captionsetup{skip=3pt} 
        \caption{Visualization of predicted frames and their corresponding prediction error maps on the UCSD Ped2, CUHK Avenue, and ShanghaiTech datasets, respectively. Normal instances are displayed in the first two rows, while abnormal instances are shown in the last two rows. Anomaly regions are marked by red bounding boxes. In the UCSD Ped2 and ShanghaiTech datasets, these two abnormal events are bicycling and driving, respectively. In the CUHK Avenue dataset, the two abnormal events are fast-running and throwing, respectively.}
        \label{fig7}
    \end{minipage}
\end{figure*}

We provide several qualitative visualization results to further illustrate the effectiveness of the proposed STNMamba for VAD tasks. Fig. \ref{fig6} shows the normality scores produced by STNMamba on four test videos from the UCSD Ped2 and CUHK Avenue datasets. As mentioned above, a frame with a higher normality score indicates more normal, while a lower means more abnormal. It can be obviously seen that the normality score remains high during normal events. However, when anomalous events such as bicycling and running occur, the score drops sharply and remains low until the anomalies subside, indicating that our STNMamba is capable of responding quickly to anomalies.

Additionally, we visualize prediction results and errors output by STNMamba on the UCSD Ped2, CUHK Avenue, and ShanghaiTech datasets, respectively. The error map is calculated by measuring the difference between the predicted frame and its ground truth, with brighter colors indicating larger prediction errors. As depicted in Fig. \ref{fig7}, the predicted frames closely resemble the corresponding ground truth during normal events, resulting in relatively small prediction errors. In contrast, the prediction results for abnormal regions are blurred, with the errors clearly highlighted in the error map. These visualization results validate the effectiveness of STNMamba in detecting abnormal events and identifying spatial locations of such anomalies.

\section{Conclusion}
In this paper, we propose the STNMamba, which explores the application of Mamba to tackle unsupervised VAD tasks for the first time, thus laying a foundation for future research. Our approach addresses the inherent challenges of spatial-temporal normality learning in VAD and the demand for computational efficiency in practical applications. Specifically, we use two carefully designed encoders to learn normality from both spatial and temporal perspectives, which are equipped with the proposed MS-VSSB and CA-VSSB to enhance the learning of spatial and temporal patterns. Additionally, we introduce an STIM that combines the developed STFB with the memory bank to model inherent spatial-temporal consistency. Extensive experiments on three benchmark datasets demonstrate that STNMamba achieves significant improvements in both performance and efficiency, highlighting its potential for practical applications in intelligent video surveillance systems. Moving forward, we intend to explore the scalability of STNMamba to other domains, such as video prediction and video action recognition, ensuring broader applicability.

\bibliographystyle{IEEEtran}
\bibliography{reference}

\end{document}